\documentclass[11pt,twoside,14pt,14pt]{article}
\usepackage[utf8]{inputenc}
\usepackage[a4paper]{geometry}
\usepackage{color}
\usepackage{textcomp}
\usepackage{amsmath}
\usepackage{amssymb}
\usepackage{graphicx}
\begin{document}

\title{Regression and Classification by Zonal Kriging}

\date{Jan 2018}

\author{Jean Serra, Jesus Angulo, B Ravi Kiran}

\maketitle
\begin{abstract}
Consider a family $Z=\{\boldsymbol{x_{i}},y_{i}$,$1\leq i\leq N\}$ of $N$ pairs of vectors $\boldsymbol{x_{i}} \in \mathbb{R}^d$ and scalars $y_{i}$ that we aim to predict for a new sample vector $\mathbf{x}_0$. Kriging models $y$ as a sum of a deterministic function $m$, a drift which depends on the point $\boldsymbol{x}$, and a random function $z$ with zero mean. The zonality hypothesis interprets $y$ as a weighted sum of $d$ random functions of a single independent variables, each of which is a kriging, with a quadratic form for the variograms drift.
    
We can therefore construct an unbiased estimator $y^{*}(\boldsymbol{x_{0}})=\sum_{i}\lambda^{i}z(\boldsymbol{x_{i}})$ de $y(\boldsymbol{x_{0}})$ with minimal variance $E[y^{*}(\boldsymbol{x_{0}})-y(\boldsymbol{x_{0}})]^{2}$, with the help of the known training set points. We give the explicitely closed form for $\lambda^{i}$ without having calculated the inverse of the matrices.
\end{abstract}

\section{Introduction}
This study follows in the line of work of \cite{Kiran-Serra-ISMM2017}.
It was motivated by work done by work done at Centre de Morphologie Mathématique in Nov 2016 \cite{Decenciere2016,AminFehri2016,RobinAlais2016}. We study two aspects of statistical learning theory, for which we propose two solutions.

\begin{enumerate}
\item The methods to construct decision trees \cite{Breiman1984} frequently constitute of partitioning the parameter space into half spaces recursively, where each binary decision optimizes various measures such as Entropy, Gini Coefficients. For possibly regionalized random variables, these methods do not take into account this intrinsic structure. Two parameter families with different different variograms shall be treated in the exactly the same way by the same algorithm.
\item Regression and classification here treated as the same problem in this decision tree framework, and we can achieve classification by doing a  threshold of the regressed variable. Though this holds in case of binary classification and we could have other cases such as the multi-class classification problem.
\end{enumerate}

We propose to take into account the regionalization of the predictors
with the application of Kriging to perform regression in the parameter space.
The theory of kriging proposed by Georges Matheron, interpolates the value attributed to each point in the space where we do not know the regionalized variable, using the neighbourhood and spatial correlation
\cite{GM_69a,GM_71a}. This not only yields as estimator but also an estimation of the variance. This is pertinent when testing hypothesis.
Thus in conclusion, we propose a method to perform regression as opposed to Breiman's method that achieves the same by partitioning the space.

\medskip{}

Kriging, like the support vector machines (SVM) \cite{Smola2002}
is formulated as an optimization of a quadratic function, but unlike (SVM) it 
uses a kernel for the extension of variances. It was applied in the area of geostatistics to cartography of mineral ores, study sea beds and 
atmosphere pressure etc. In the present case, its use in multivariate 
statistics differs from geostatistics. Firstly the space is no more geographical, but a set of parameters. We consider a family $Z=\{\boldsymbol{x}_{i},y_{i}$,$1\leq i\leq N\}$
of $N$ pairs of parameters $\boldsymbol{x_{i}}$ and measures $y_{i}$.

The $\boldsymbol{x_{i}}$s are inputs and the $y(\boldsymbol{x}_{i})=y_{i}$
the outputs. Every value $\boldsymbol{x_{i}}$ is a vector consisting of
$d$ scalar parameters $x_{i,1},x_{i,2}...x_{i,s},...x_{i,d}$
each coming from $\mathbb{R}^1$,
and we would like to predict the measure $y$ at a new test point 
$\boldsymbol{x}_{0}=(x_{0,1},x_{0,2},...x_{0,s,...}x_{0,d})$ not in
the family $Z$. Once the estimation is done using the training set, 
we use another sample set, test set to evaluate the performance of the estimation.

The second difference lies in the size of the training set.
The order of $N$ and $d$ are variable by are sometimes quite large. A family of an average size allows for 1000s of parameters to be associated to 1000s of samples.In other situations, $N$ could be millions. But the kriging estimator at a point of the space requires that we solve a linear system of $N$ variables. Could we avoid inverting a matrix now that is around $10^{3}\times10^{3}$, i.e. with $10^{6}\times10^{6}$ elements ?

\paragraph{Motivation}
This study's primary goal is to express the estimator
$y^{*}(\boldsymbol{x}$) over all points $\mathbf{x} \in  \mathbb{R}^d$
without inverting a matrix of $N\times N$, while using efficient algorithms. Following this goal we shall decompose
the estimators over $\mathbb{R}^{d}$ to scalar uni-dimensional estimators with independent scalar variables. The second objective consists in finding
the solution for classification using vector regressions 
following similar decomposition.

\section{Reminder on kriging}
Kriging is an optimal prediction of $y$, defined on a metric space $D$.
To estimate $y$ at new test point $\mathbf{x}_0$, we use the linear function $z(x_{i})$ for $x_{i} \in D$ while taking into account the interdependence of the training samples.

The idea is to interpret $y$ as a random function with the mean value $m$
and the residue $z$ :

\begin{equation}
\left\{ \begin{array}{l}
y(x)=m(x)+z(x)=a_{0}+a_{1}x+a_{2}x^{2}+z(x)\\[2ex]
E[y(x)]=m(x)\quad;\quad E[z(x)]=0
\end{array}\right.\label{eq:base_KU}
\end{equation}

The mean, or the drift, $m$ is a deterministic function, though one that depends on the point $x$, while $z$ is a random stationary function with zero mean and covariance $\sigma$. We say that the estimated a deterministic value, but with unknown $m(\mathbf{x})$, and we krige the random values $y(\mathbf{x}$ and $z(\mathbf{x})$ (and not their mean values).
The first objective remains the kriging of $y$, but one the objective is to estimate the drift and krige the residue.The functions $m$
and $\sigma$ are not arbitrary, the terms of the drift should be linearly independent, and the covariance positive definite.

In the present case four enumerations appear, with one part of the
variation from $1$ to $d$ parameters, running index $s$,
and then the cell, from $1$ to $N$ of samples, running index $i$
and finally the coefficients of the drift, from $0$ to $3,$ running index $u$. Now, for the problem of classification that will touch upon later, it is also required to take into account the enumeration over classes. 
However, the four-indexes lead to unreadable equations, we successively treat
the estimation of the drift $m$, and then the kriging to estimate $y$. These two steps are done in line with scalar regression; while when the classification shall be studied shall be interpreted as a vector Kriging.

Regarding notations, Matheron uses Einstein convention for implicit sums,
while the more recent book by Chilès \& Deflfiner avoids to write a more compact  matrice representation of the linear systems, which is usually the norm in the machine learning community. In this study we keep the old notations.

\subsection{Estimation of the drift in $\boldsymbol{x_{0}}$}

Given the space $\mathbb{R}^{d}$ of parameters, we would like to estimate
the for $\mathbf{x}_{0}$ the drift $m(\boldsymbol{x_{0}})$, this is introduced in the expression (\ref{eq:base_KU}). First we give a model for the drift assumed to be quadratic :

\begin{equation}
m(\boldsymbol{x})=a_{0}+a_{1}\boldsymbol{x}+a_{2}\boldsymbol{x}^{2},\label{eq:modele_derive}
\end{equation}
We then form the linear composition :
\begin{equation}
m^{*}(\boldsymbol{x_{0}})=\sum_{i}\mu^{i}y(\boldsymbol{x_{i}})\label{eq:estimateur_moyenne}
\end{equation}
and we require the $\mu^{i}$ be universal, that is whatever $a_{u}$, $\:0\leq u\leq2$, no bias is introduced, i.e.: 
\begin{equation}
E[m^{*}(\boldsymbol{x_{0}})]=m(\boldsymbol{x_{0}}).\label{eq:non_biais_m}
\end{equation}
This gives us three constraints
\begin{equation}
\sum_{i}\mu^{i}=1\qquad\sum_{i}\mu^{i}\boldsymbol{x_{i}}=\boldsymbol{x_{0}}\qquad\sum_{i}\mu^{i}\boldsymbol{x_{i}}^{2}=\boldsymbol{x_{0}}^{2}.
\label{eq:universality_constraint}
\end{equation}
The estimation of variance of $m$ is given by: 
\[
E[m(\boldsymbol{x_{0}})-m^{*}(\boldsymbol{x_{0}})]^{2}=E\left\{ m(\boldsymbol{x_{0}})-\sum_{i}\mu^{i}(z(\boldsymbol{x_{i}})+m(\boldsymbol{x_{i}})\right\} ^{2}
\]
which when takes account of the universality conditions in (\ref{eq:universality_constraint})
simplifies to
\begin{equation}
E[m(\boldsymbol{x_{0}})-m^{*}(\boldsymbol{x_{0}})]^{2}=\sum_{j}\sum_{i}\mu^{j}\mu^{i}\sigma(i,j)\label{eq:variance_derive}
\end{equation}
where $\sigma(i,j)=E[z(\boldsymbol{x_{i}})z(\boldsymbol{x_{j}})]$ is
the covariance between the residues $z$($\boldsymbol{x_{i}})$ and $z$($\boldsymbol{x_{j}})$. We obtain the coefficients $\mu^{i}$ by minimizing this estimation variance which accounts for the three constraints in (\ref{eq:universality_constraint}).
The classical Lagrange approach yields three  multipliers $\rho_{u}$
and leads to the system of $N+3$ equations :

\begin{equation}
\left\{ \begin{array}{l}
\sum_{j}\mu^{j}\sigma(i,j)=\rho_{0}+\rho_{1}\boldsymbol{x}+\rho_{2}(\boldsymbol{x})^{2}\qquad\quad1\leq i\leq N\\[2ex]
\sum_{i}\mu^{i}=1\qquad\sum_{i}\mu^{i}\boldsymbol{x_{i}}=\boldsymbol{x_{0}}\qquad\sum_{i}\mu^{i}\boldsymbol{x_{i}}^{2}=\boldsymbol{x_{0}}^{2}
\end{array}\right.\label{eq:system_derive}
\end{equation}

The $\mu^{i}$s are the coefficients of the optimal estimator (\ref{eq:estimateur_moyenne}), we obtain the estimation variance of the estimate
$m$ by multiplying with $\mu^{i}$ the first relation of the system,
and then we sum over $i$ and use the second, which finally gives :
\begin{equation}
E[m(\boldsymbol{x_{0}})-m^{*}(\boldsymbol{x_{0}})]^{2}=\rho_{0}+\rho_{1}\boldsymbol{x_{0}}+\rho_{2}(\boldsymbol{x_{0}})^{2}.\label{eq:variance_krigeage_y}
\end{equation}

\subsubsection{Estimation of coefficients of the drift}

We can refine the estimate of the drift while detailing the estimations
of the three coefficients $a_{0},a_{1},a_{2}$, since we have
\[
m^{*}(\boldsymbol{x})=a_{0}^{*}+a_{1}^{*}\boldsymbol{x}+a_{2}^{*}\boldsymbol{x}^{2}
\]
the method consists of representing each of these three terms  $a_{0}^{*},a_{1}^{*},a_{2}^{*}$
as the weighted sums $y(\boldsymbol{x_{i}})$ and then minimizing their variances
following the Lagrangian method as usual. We do not indicate
the steps, which can be found in \cite{GM_71a}. Though we give for the case of zonal kriging.

The estimators are useful for reducing the drift and the first and second terms.
So the constant drift estimator $m(\boldsymbol{x})=a_{0}$
is $m^{*}(\boldsymbol{x})=a_{0}^{*}$, and that of linear drift
$m(\boldsymbol{x})=a_{0}+a_{1}\boldsymbol{x}$ is $m^{*}(\boldsymbol{x})=a_{0}^{*}+a_{1}^{*}\boldsymbol{x}.$

\subsection{Kriging of $y$ in $\boldsymbol{x_{0}}$}

We follow the same procedure as before. To krige $y$ in $\boldsymbol{x_{0}}$
we form an linear estimator of $y(\boldsymbol{x_{i}})$ , i.e.
\begin{equation}
y^{*}(\boldsymbol{x_{0}})=\sum\lambda^{i}y(\boldsymbol{x_{i}})\label{eq:krigeage_y-1}
\end{equation}
and we minimize the variance of the error $E[y(\boldsymbol{x_{0}})-y^{*}(\boldsymbol{x_{0}})]^{2}$
conditionally on the three constraints (\ref{eq:universality_constraint}),
and introduce three Lagrange multipliers $\tau_{u}$, which gives $N+3$ equations :
\begin{equation}
\left\{ \begin{array}{l}
\sum_{j}\lambda^{j}\sigma(i,j)=\sigma(i,0)+\tau_{0}+\tau_{1}\boldsymbol{x}+\tau_{2}(\boldsymbol{x})^{2}\qquad\quad1\leq i\leq N\\[2ex]
\sum_{i}\lambda^{i}=1\qquad\sum_{i}\lambda^{i}\boldsymbol{x_{i}}=\boldsymbol{x_{0}}\qquad\sum_{i}\lambda^{i}\boldsymbol{x_{i}}^{2}=\boldsymbol{x_{0}}^{2}
\end{array}\right.\label{eq:system_krigeage-1}
\end{equation}
and the minimal variance
\[
E[y(\boldsymbol{x_{0}})-y^{*}(\boldsymbol{x_{0}})]^{2}=\sigma^{2}(y(\boldsymbol{x_{0}}))-\lambda^{i}\sigma(i,0)+\tau_{0}+\tau_{1}\boldsymbol{x_{0}}+\tau_{2}\boldsymbol{x_{0}}{}^{2}.
\]
Instead of attacking the front of the system (\ref{eq:system_krigeage-1}),
in general its simpler to combine estimation of the drift with the kriging of the resisdue at $\boldsymbol{x_{0}}$. For this we use the linear estimator :
\begin{equation}
z^{*}(\boldsymbol{x_{0}})=\sum_{i}\nu^{i}z(\boldsymbol{x_{i}})\label{eq:estimateur_simple}
\end{equation}
of $z(\boldsymbol{x_{0}}$) from the $z(\boldsymbol{x_{i}})$.
In this case there is no universality constraint, and is sufficient to  minimize the variance of the error $E[z^{*}(\boldsymbol{x_{0}})-z(\boldsymbol{x_{0}})]^{2}$
to obtain the system 
\begin{equation}
\sum_{j}\nu^{j}\sigma(i,j)=\sigma(i,0),\label{eq:system_residus}
\end{equation}
of $N$ equations and $N$ unknowns which gives the coefficients
$\nu^{i}$. The corresponding kriging variance is 
\begin{equation}
E[z^{*}(\boldsymbol{x_{0}})-z(\boldsymbol{x_{0}})]^{2}=\sigma(0,0)+\sum_{i}\nu^{i}\sigma(i,0)\label{eq:variance_residus}
\end{equation}

However, we cannot directly construct the estimator (\ref{eq:estimateur_simple}),
since we only know, for each $\boldsymbol{x_{i}}$, the value of $y(\boldsymbol{x_{i}})$
and the optimal estimate of the drift $m^{*}(\boldsymbol{x}_{i}$).
But we can use the theorem of kriging additivity \cite{GM_71a}, which states that we can   krige the residues $y(\boldsymbol{x_{i}})-m^{*}(\boldsymbol{x_{i}})$
as if it were the true residues. Consequently the estimator 
\begin{equation}
z^{**}(\boldsymbol{x_{0}})=\sum_{i}\nu^{i}[y(\boldsymbol{x_{i}})-m^{*}(\boldsymbol{x_{i}})]\label{eq:estimateur_residus}
\end{equation}
is optimal when the $\nu^{i}$s are solutions to the system (\ref{eq:system_residus})
and evaluates to $z^{*}(\boldsymbol{x_{0}})$, with a 
 kriging variance
\begin{equation}
E[z^{*}(\boldsymbol{x_{0}})-z(\boldsymbol{x_{0}})]^{2}=\sigma(0,0)+\sum_{i}\nu^{i}\sigma(i,0).\label{eq:variance_krigeage_residus}
\end{equation}

The kriging $y^{*}(\boldsymbol{x_{0}})$ of $y$ at $\boldsymbol{x_{0}}$ and its variance is therefore
given by: 
\begin{equation}
y^{*}(\boldsymbol{x_{0}})=\sum_{i}[(\mu^{i}+\nu^{i})y(\boldsymbol{x_{i}})-\nu^{i}m^{*}(\boldsymbol{x_{i}})]=\sum_{i}\lambda^{i}y(\boldsymbol{x_{i}})\label{eq:system_krigeage}
\end{equation}

and 
\[
E[y(\boldsymbol{x_{0}})-y^{*}(\boldsymbol{x_{0}})]^{2}=\rho_{0}+\rho_{1}\boldsymbol{x_{0}}+\rho_{2}(\boldsymbol{x_{0}})^{2}+\sigma(0,0)+\sum_{i}\nu^{i}\sigma(i,0)
\]
where these $\lambda^{i}$ are the weights attributed to $y(\boldsymbol{x_{i}})$ in the equation (\ref{eq:krigeage_y-1}), but now that are calculated easily from the $\mu^{i}$s and the $\nu^{i}$s. The two systems
(\ref{eq:system_derive}) et (\ref{eq:system_residus}) of estimation
of the drift and the residue are regular, by convexity of the variance.
Each admits a unique solution.

\subsection{Covariance or Variogram?}
We have modeled the residue $z$ as a random stationary function with zero 
mean and covariance $\sigma(x,x')$, which implies a finite variance $\sigma(0,0)$.
Now this model of residue does not always work in real life. We observe in 
geosciences certain phenomena like the distribution of minerals in mines, 
or predicting rainfalls, where the capacity of dispersion appears infinite.
Larger the period or the region we study, larger their variance, without 
the appearance of any kind of limit in the order of values that are measured. The covariance, whose value at the origin is the variance, is therefore undefined, and we must avoid it in the formalism. Fortunately, the variance of the increments $y(x)-y(x')$ is still finite for finite distances between $x$ and $x'$, 
and the tool which expresses it is the variogram $\gamma(x,x')$ 
\begin{equation}
\gamma(x,x')=\frac{E[y(x)-y(x')]^{2}}{2}=\sigma(00)-\sigma(x,x').\label{eq:variogramme-covariance}
\end{equation}

Moving from covariance to variogram in the problem of kriging
is equivalent to replacing the model (\ref{eq:base_KU}) of the
function $y$ by the following: 
\begin{equation}
\left\{ \begin{array}{l}
y(x)=m(x)+z(x)=a_{1}x+a_{2}x^{2}+z(x)\\[2ex]
E[z(x)]=a_{0}\quad;\quad E[y(x)]=m(x)+a_{0}
\end{array}\right.\label{eq:base_ku_vario}
\end{equation}
where the mean does not have the constant term anymore and the residue
$z$ admits $\gamma(x,x')$ for the variogram. When we replace the $\sigma(x,x')$ by $\sigma(00)-\gamma(x,x')$ in System (\ref{eq:system_derive}) and in Equation (\ref{eq:system_residus}),  then the $\sigma(00)$ 
get eliminated: we find the same solutions for $\mu^{i}$ as before, and also for $\rho_{u}$ up to a change of sign. Concerning the residue
one should assure that the two measures of the kriging (\ref{eq:estimateur_simple}) 
have the same expectation, which implies a condition of universality on 
the $\nu^{i}$ and leads to a system :
\begin{equation}
\left\{ \begin{array}{l}
\sum_{j}\nu^{j}\gamma(i,j)=\gamma(i,0)-\tau_{0}\qquad\quad1\leq i\leq N\\[2ex]
\sum_{i}\nu^{i}=1\qquad
\end{array}\right.\label{eq:system_ku_vario}
\end{equation}
We obtain the kriging for $y$ at $\boldsymbol{x_{0}}$ by applying the theorem
of additivity to the sum

\begin{equation}
y^{*}(\boldsymbol{x_{0}})=a_{1}^{*}\boldsymbol{x_{0}}+a_{2}^{*}
\boldsymbol{x_{0}}{}^{2}+z^{*}(\boldsymbol{x_{0}})
\label{eq:krigeage_par_variogramme}
\end{equation}
where the term $a_{0}$ has disappeared, which shows that kriging
involves increments only. The associated variance is
\[
E[y(\boldsymbol{x_{0}})-y^{*}(\boldsymbol{x_{0}})]^{2}=\sum_{i}\lambda^{i}\gamma(i,0)+\rho_{0}+\rho_{1}\boldsymbol{x_{0}}+\rho_{2}(\boldsymbol{x_{0}})^{2}.
\]

The term $a_{0}$ is useless for the kriging of $y,$ and does not have any significance when the neighbourhood is the whole space $\mathbb{R}^d$. In retrospect it becomes meaningful when the data $\boldsymbol{x_{i}}$
occupy a bounded portion of the space, or when we perform  slipping neighbourhoods, with a bounded support centered at $\boldsymbol{x_{0}}$, to
map the drift of $y$ . The estimator $a_{0}^{*}$ of $a_{0}$ which results from these equations (\ref{eq:system_derive}),
in their variogram version, just provides the drift $m^{*}(\boldsymbol{x_{0}})$ at $\boldsymbol{x_{0}}$, since the terms at $x$ and at $x^{2}$ cancel at the origin.

\section{Kriging in the presence of zonal anisotropy}

The two systems (\ref{eq:system_derive}) \& (\ref{eq:system_residus})
of equations (\ref{eq:system_krigeage}) respond to the question of  
optimal estimation of $y$ \& of $m$, but at the price of inversion of 
matrices of size $(N+3)\times(N+3)$ and it requires to work in the space $\mathbb{R^{\textrm{\ensuremath{d}}}}$. These usually pose computational 
constraints for large training samples set that now we shall try to solve 
using zonal anisotropy. This will allow us to decompose the multidimensional 
kriging of $y$ into $d$ univariate krigings, such that each parameter 
generates separately a term of the estimator of
$y$ at point $\boldsymbol{x}_{0}=x_{01},...x_{0s,...}x_{0d}$.

\subsection{Zonality hypothesis}
In geostatistics, zonality is a form of anistropy where the regionalized
variable $y(x_{1},x_{2,}x_{3})$ of the physical space with 3 dimensions
does varies only in one direction \cite{GM_71a,Chiles-Delfiner_2015-1,Wackernagel}.
The classical example is of that of strata in vertical layers of sediments,
which vary very little horizontally. The zonality is only in the vertical direction, 
but we can combine with other anisotropies, like for example two other zonalities in the 
horizontal directions.

When we place ourselves in the space $\mathbb{R}^d$ of parameters
the hypothesis zonal anisotropy is formulated by taking interpreting
$y$, in each neighbourhood where the kriging will be applied next, 
as the weighted sum of $d$ random functions on $\mathbb{R}^d$
of only one parameter $s$ each,
and whose cross variations are independent:
\begin{equation}
y(\boldsymbol{x})=w_{1}y_{1}(x_{1})+...w_{s}y_{s}(x_{s})+...w_{d}y_{d}(x_{d}) 
\label{eq:zonality}
\end{equation}
The hypothesis of linearity of the components is a simple extension
of the linearity of regression model in $\mathbb{R}^1$. The independence
of the components is a strong assumption and which will be discussed in section \ref{sec:Modes-op}. 
 When we pass on to kriging every $y_{s}$ is modeled as a sum of a drift and a residue with
\[
y_{s}(x_{s})=m_{s}(x_{s})+z_{s}(x_{s})
\]
where
\[
y(\boldsymbol{x})=\sum_{s}w_{s}m_{s}(x_{s})+\sum_{s}w_{s}z_{s}(x_{s})=m(\boldsymbol{x})+z(\boldsymbol{x}).
\]
Let us see now how the zonality hypothesis simplifies the kriging of residue
$z(\boldsymbol{x_{0}})$ and of $y(\boldsymbol{x_{0}})$ at point $\boldsymbol{x_{0}}$.

\subsection{Zonal kriging of the residue}
The zonal residue $z$ can be written as sum of $d$ random functions
$z_{s}$, stationary and independent, and each with zero mean: 
\begin{equation}
z(\boldsymbol{x})=\sum_{s}w_{s}z_{s}(x_{s})\qquad
\hfill\boldsymbol{x}=(x_{1}...,x_{s}...,x_{d}).
\label{eq:residu_zonal}
\end{equation}

Let us note that $\sigma_{s}(i,j)$ the covariance of $z_{s}$ at $s$-abscissas
$x_{i,s}$ \& $x_{j,s}$ \& $\sigma(i,j)$ and that of values of $z$,
or of $y,$ at points $\boldsymbol{x_{i}}$ and $\boldsymbol{x_{j}}$, i.e.
\[
\mathbb{E}[z_{s}(x_{i,s})z_{s}(x_{j,s})]=\sigma_{s}(i,j)\quad\text{and}\quad \mathbb{E}[z(\boldsymbol{\boldsymbol{x}_{i}})z(\boldsymbol{x_{j}})]=
\mathbb{E}[y(\boldsymbol{\boldsymbol{x}_{i}})y(\boldsymbol{x_{j}})]=\sigma(i,j)
\]

The cross independence of functions $z_{s}$ implies that
\begin{equation}
E[z_{s}(x_{i,s})z_{t}(x_{j,t})]=0\qquad\forall s,t\in[1...d]\quad s\neq t
\label{eq:covar_zonale_ij}
\end{equation}
where
\begin{equation}
\sigma(i,j)=\sum_{s}w_{s}^{2}\sigma_{s}(i,j)
\label{eq:covar_zonal_total}
\end{equation}
In the same way, the covariance between the function $z$ at point $\boldsymbol{x_{i}}$
and the function $w_{s}z_{s}$ at point $x_{j,s}$, when accounted for zonal anisotropy, is equal to 

\begin{equation}
\mathbb{E}[z(\boldsymbol{x_{i}})w_{s}z_{s}(x_{j,s})]=
\mathbb{E}[w_{s}z_{s}(x_{i,s})w_{s}z_{s}(x_{j,s})]=w_{s}^{2}\sigma_{s}(i,j)
\label{eq:covar_zonale_i_total}
\end{equation}.

The coefficients of kriging
\begin{equation}
z^{*}(\boldsymbol{x_{0}})=\sum_{i}\nu^{i}z(\boldsymbol{x_{i}})\label{eq:krigeage_classique_residus}
\end{equation}
of $z(\boldsymbol{x_{0}})$ with the help of points $z(\boldsymbol{x_{i}})$
satisfies the system of equations (\ref{eq:system_residus}), i.e.
$\sum_{j}\nu^{j}\sigma(i,j)=\sigma(i,0)$. It is not sure that estimator
$z^{*}(\boldsymbol{x_{0}})$ thus obtained, even if optimal, satisfies
the zonal decomposition (\ref{eq:residu_zonal}). However, we can decompose the value
 $z^{*}(\boldsymbol{x_{0}})$ at point $x_{0}$ into zonal terms:
\[
z^{*}(\boldsymbol{x_{0}})=\sum_{j}\nu^{j}\sum_{s}w_{s}z_{s}(x_{i,s})
\]

The corresponding estimation variance can then be written, given the expressions
of covariances (\ref{eq:covar_zonal_total}) and (\ref{eq:covar_zonale_i_total})
\[
\mathbb{E}[z^{*}(\boldsymbol{x_{0}})-z(\boldsymbol{x_{0}})]^{2}=
\mathbb{E}[z(\boldsymbol{x_{0}})]^{2}-2\sum_{i}\nu^{i}\sum_{s}w_{s}^{2}\sigma_{s}(i,0)+\sum_{j}\sum_{i}\nu^{j}\nu^{i}\sum_{s}w_{s}^{2}\sigma_{s}(i,j).
\]

This variance is minimal when all its derivatives at $\nu^{i}$
are null, which leads to the system at $ N $ equations:
\begin{equation}
\sum_{j}\nu^{j}\sum_{s}w_{s}^{2}\sigma_{s}(i,j)=\sum_{s}w_{s}^{2}\sigma_{s}(i,0).\qquad\quad1\leq i\leq N\label{eq:krigeage_zonal}
\end{equation}

On the other hand, one can write the system for kriging of the component  $z_{s}$ of $z$ at point $x_{0,s}$ with the help of components $z_{s}(x_{i,s})$
at known points $x_{i,s}$. It suffices to form the estimator $z_{s}^{*}(x_{0,s})$:
\begin{equation}
z_{s}^{*}(x_{0,s})=\sum_{i}\nu_{s}^{i}z_{s}(x_{i,s})\label{eq:estim_zs}
\end{equation}
of estimation variance
\begin{equation}
E[z_{s}^{*}(x_{0,s})-z_{s}(x_{0,s})]^{2}\label{eq:var_estim_zs}
\end{equation}
whose minimization leads to $N$ equations (\ref{eq:system_residus}):
\begin{equation}
\sum_{j}\nu_{s}^{j}\sigma_{s}(i,j)=\sigma_{s}(i,0)\,,\qquad\quad1\leq i\leq N
\label{eq:krigeage_composante_s}
\end{equation}

The kriging $z^{*}(\boldsymbol{x_{0}})$ and those of the various components $z_{s}^{*}(x_{0,s})$
are linked. The sum in $s$ of the estimation variances  (\ref{eq:var_estim_zs}),
weighted by the coefficients $w_{s}^{2}$, is equal to
\begin{equation}
w_{s}^{2}\sum_{s}E[z_{s}^{*}(x_{0,s})-z_{s}(x_{0,s})]^{2}=E\left\{ \sum_{s}w_{s}^{2}[z_{s}^{*}(x_{0,s})-z_{s}(x_{0,s})]\right\} ^{2}\label{eq:somme_s_variances}
\end{equation}
because the increments of the components $z_{s}$ are independent
between them. When the variances of the right member are minimal,
that of the left member also become minimal, which implies

\begin{equation}
z^{*}(\boldsymbol{x_{0}})=\sum_{s}z_{s}^{*}(x_{0,s}).\label{eq:decomposition}
\end{equation}

Since we do not know the terms $z_{s}(x_{i, s})$ that intervene
in the sum (\ref{eq:estim_zs}), the decomposition (\ref{eq:decomposition})
is not sufficient to determine the coefficients $\nu^{j}$ of
$Z^{*}(\boldsymbol{x_0})$. The set of $z_{s}(x_{i, s})$ contains
moreover, more information than their sums $z(x_{i})$. However,
if we add the $d$ systems (\ref{eq:krigeage_composante_s}),
we obtain: 
\begin{equation}
\sum_{j}\sum_{s}\nu_{s}^{j}w_{s}^{2}\sigma_{s}(i,j)=\sum_{s}w_{s}^{2}\sigma_{s}(i,0)\label{eq:somme_krigeages_s}
\end{equation}

So we see that if we set $i$ to: 
\begin{equation}
\nu^{j}(i)=\frac{\sum_{s}\nu_{s}^{j}w_{s}^{2}\sigma_{s}(i,j)}{\sum_{s}w_{s}^{2}\sigma_{s}(i,j))}\qquad
\label{eq:krigeage_residus_1_D},
\end{equation}
then the $\bar{\nu^{j}(i)}$ satisfies the $i$ parameter equation
system (\ref{eq:krigeage_zonal}), but not necessarily the others.
However, we can no longer depend on $i$ by introducing the term
$\tau_{s}(j) = \sum_{i} w_{s}^{2} \sigma_{s}(i, j)$ which appears when
sum in $i$ the equations (\ref{eq:krigeage_residus_1_D}). it
leads to the new coefficient $\nu^{*j}$
\begin{equation}
\nu^{*j}=\frac{\sum_{s}\nu_{s}^{j}\tau_{s}(j)}{\sum_{s}\tau_{s}(j)}.\label{eq:ersatz_poids_residus}
\end{equation}
which can be interpreted as a weighted average in $i$ of $\bar{\nu^{j}(i)}$.
These coefficients suggest the new estimator $z^{**}(\boldsymbol{x_ {0}})$:
\begin{equation}
z^{**}(\boldsymbol{x_{0}})==\sum_{i}\nu^{*i}z(\boldsymbol{x_{i}})
\label{eq:ersatz_estimateur_residus}
\end{equation}
whose coefficients $\nu^{*i},1\leq i\leq d$ are not those anymore
of the estimator (\ref{eq:krigeage_classique_residus}). On the other hand,
it satisfies the condition of zonality and its coefficients put in
set of 1-D krigings only. So, the knowledge of
coefficients relating to 1-D residues, and their covariance, 
are sufficient to determine the weights to estimate
residues in space $\mathbb{R^{\textrm{\ensuremath{d}}}}$.

\subsection{Estimation of the drift}

Let us first treat the case where the drift is reduced to a constant term
$A_{0}$. This allows to start with simpler notation, and which a situation is very instructive by itself. We
thus suppose that $ y $ is a stationary random function but
we do not know the average $ a_ {0} $. As in the theory
general, we form the estimator

\[
m^{*}(\boldsymbol{x_{0}})=\sum_{i}\mu^{i}y(\boldsymbol{x_{i}}),
\]
which is not biased when $a_{0} = m(\boldsymbol{x_{0}}) = \mathbb{E}[m ^ {*} (\boldsymbol{x_ {0}})]$
, i.e. when $\sum_{i} \mu^{i} = 1$. The zonal decomposition of $m^{*}(\boldsymbol{x_ {0}}$
is written as:

\begin{equation}
m^{*}(\boldsymbol{x_{0}})=\sum_{i}\mu^{i}\sum_{s}w_{s}y_{s}(x_{i,s})
\label{eq:moyenne_derive}
\end{equation}

Every $y_{s}$ is a stationary random function of average $a_{0, s}$ with 

\begin{equation}
a_{0}=\sum_{s}w_{s}a_{0,s}.
\label{eq:a_zero}
\end{equation}

The minimization for system of equations (\ref{eq:system_derive})
reduces to

\begin{equation}
\left\{ \begin{array}{l}
\sum_{j}\mu^{j}\sum_{s}w_{s}^{2}\sigma_{s}(i,j)=\rho_{0}\qquad\quad1\leq i\leq N\\[2ex]
\sum_{j}\mu^{j}=1
\end{array}\right.\label{eq:system_KU-1}
\end{equation}

where $\rho_{0}$ is the Lagrange parameter and where the covariance between
the $y(\boldsymbol{x_ {i}})$,  and $y(\boldsymbol{x_ {j}})$ is $\sigma(i, j)$.
Similarly, we form on each axis $ s $ the estimator

\begin{equation}
m_{s}^{*}(x_{0,s})=\sum_{i}\mu_{s}^{i}y_{s}(x_{i,s})\label{eq:estimateurs_ms}
\end{equation}
of Lagrange minimization

\begin{equation}
\left\{ \begin{array}{l}
\sum_{j}\mu_{s}^{j}\sigma_{s}(i,j)=\rho_{0,s}\qquad\quad1\leq i\leq N\\[2ex]
\sum_{j}\mu_{s}^{j}=1.
\end{array}\right.\label{eq:system_KU_des_s}
\end{equation}

By summation in $ s $ we find for the first equations
\[
\sum_{s}\sum_{j}\mu_{s}^{j}w_{s}^{2}\sigma_{s}(i,j)=\sum_{s}w_{s}^{2}\rho_{0,s}\qquad1\leq i\leq N
\]

If we put
\begin{equation}
\rho_{0}=\sum_{s}w_{s}^{2}\rho_{0,s}\label{eq:valeur_de_ro}
\end{equation}
and if we suppose that the systems (\ref{eq:system_KU_des_s})
for all $s$, then we deduce the $\mu^{j}$s: 

\begin{equation}
\mu^{j}=\frac{\sum_{s}\mu_{s}^{j}w_{s}^{2}\sigma_{s}(i,j)}{\sum_{s}w_{s}^{2}\sigma_{s}(i,j))}
=\frac{\sum_{s}\mu_{s}^{j}\tau_{s}(j)}{\sum_{s}\tau_{s}(j)}
\label{eq:mu_j}
\end{equation}
i.e. the solutions of the first equations (\ref{eq:system_KU-1}). As the
estimates (\ref{eq:estimateurs_ms}) are not biased, we find, by taking the mathematical expectations in (\ref{eq:moyenne_derive})
\[
E[m^{*}(\boldsymbol{x_{0}})]=\sum_{i}\mu^{i}\sum_{s}w_{s}E[y_{x}(x_{i,s})]=\sum_{i}\mu^{i}\sum_{s}w_{s}a_{0,s}=a_{0}\sum_{j}\mu^{j}
\]

The sum of the $\mu^{j}$ of (\ref{eq:mu_j}) is therefore 1 so that they are the solution of the system (\ref{eq:system_KU-1}).

If now the drift is quadratic, the same reasoning shows
that the coefficients $\mu^{i}$ are still expressed according to the
$\mu_{s}^{j}$ by the equations (\ref{eq:mu_j}), but the $\mu_{s}^{j}$
are no longer the same as for a constant drift. When the
coefficients $\mu_{s}^{j}$ one-dimensional drifts verify
the three conditions (\ref{eq:universality_constraint}),
then the $\mu^{i}$ still satisfy them. The approach extends
besides any drift of the type $m = \sum a_{p} f^{p}$ where $ f^{p}$
are linearly independent functions (to ensure regularity
systems of equations).
\subsection{Zonal kriging of $y$}

As for the kriging $y^{*}(\boldsymbol{x_{0}})$ of $y(\boldsymbol{x_{0}})$
it suffices to apply the additivity theorem, which gives :
\begin{equation}
y^{*}(\boldsymbol{x_{0}})=m^{*}(\boldsymbol{x_{0}})+
\nu^{i}[y(\boldsymbol{x_{i}})-m^{*}(\boldsymbol{x_{i}})]
\label{eq:krigeage_y}
\end{equation}
where the coefficients of the estimators $m^{*}$ are those of the equations
(\ref{eq:mu_j}). This result is fundamental. Each coefficient
$\nu^{i}$ (resp. $\mu^{i}$) appears in (\ref{eq:ersatz_poids_residus})
as the average of $\nu_{s}^{j}$ (resp. $\mu_{s}^{i}$) are 1-D,
weighted by the covariances $\tau(i, j)$. We further determine 
these $\nu_{s}^{j}$ and $\mu_{s}^{i}$ by formal computation in $\mathbb{R}^1$,
valid even for a very large $N$. It requires a covariance model
$\sigma_{s}(i, j)$ or associated variogram $\gamma_{s}(i, j)$.
We will choose the linear variogram, with or without a nugget effect,
because it's the most common. On the other hand, we do not need to estimate
the zonal decomposition of $y$ into its components $y_{s} $ of (\ref{eq:zonality}),
which is a considerable advantage.

We have introduced weights $w_{s}$ into the expression (\ref{eq:zonality})
the zonality of $y$ since the $s$ parameters are not physically
homogeneous. One can designate a temperature, another a chemical 
content, a third an orientation, etc. Since their variation
depends on the choice of their units, the weights $w_{s}$ allow
to take this into account. They can be used for example to 
calibrate the parameters $s$ between them by requiring them to have the same variance.

\subsection{Why the zonal hypothesis?}
As in decision trees, where the euclidean space $\mathbb{R}^{3}$
is replaced by a product of d-$\mathbb{R}^1$, spaces
the hypothesis of zonality represents four significant advantages: 

\begin{enumerate}
\item The dimensions of the geographical space are each physically equivalent, 
whereas the same can not be said of a parameter space,
even if they are eventually correlated. In more technical terms,
let us say the space $\mathbb{R}^d$ of parameters, 
the only balls that have a physical sense are the rectangle  parallelepipeds 
parallel to the axes.
It's only relative to these distances we can build variograms
and krige these points of $\mathbb{R}^d$ which is in accordance 
with the zonal decomposition.

\item Random Forests manipulate subspaces of parameters drawn at random from
$\mathbb{R}^d$ and that lead to decision subtrees. Zonal independence
allows the kriging of each parameter for any subspace.

\item In the earth sciences, when the number of samples $N$ is
too high, it is not any more trivial to solve the kriging system.
The difficulty is overcome by means of sliding neighborhoods, which reduce
locally the volume of data. But the notion of neighborhood implies
that the function $y$ is sampled fairly homogeneously
throughout the space. But this pseudo geographical regularity makes little
sense in a parameter space: the $s$ coordinate to be predicted at a point
can be very close to that of a known point but very
distant from the coordinate $t$ of the same known point. By making
systems of one-dimensional equations, zonal kriging allows
moving window neighborhoods since they are then independent of an axis
to the other.

\item On the other hand, if the zonality drastically reduces dimensions of
parameter spaces, it does not affect the number of samples.
On the other hand, we manage to formally solve the systems of equations
kriging in some cases, because they relate to $\mathbb{R}^1$,
as we will see now.
\end{enumerate}

\section{1-D regression by kriging }
In $\mathbb{R}^1$ in fact, we know to solve the equations of kriging for the standard variograms. G. Matheron calculated for a certain number\footnote{The texts are available at the website http://cg.ensmp.fr/bibliotheque/cgi-bin/public/bibli\_index.cgi }
for the universal Kriging \cite{GM_69a}\textit{,} \textit{La
théorie des variables régionalisées et ses application}s \cite{GM_71a},
and the \textit{Note de géostatistique n\textdegree{} 106, Exercices
sur le krigeage universel} \cite{GM_70a}. We review and consider one of these cases in this section, without proofs, and with discrete calculations.

\begin{figure}
\begin{centering}
\includegraphics[scale=0.5]{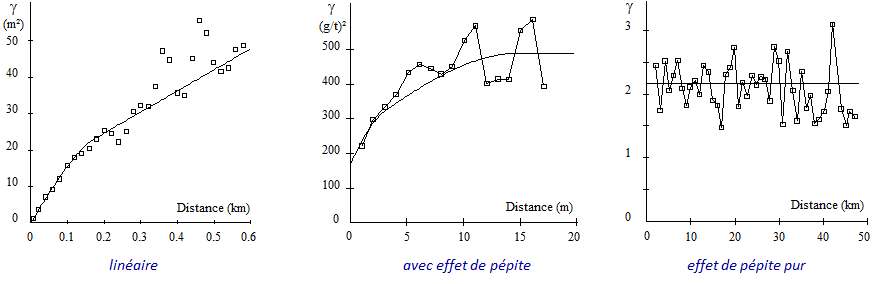} 
\par\end{centering}
\caption{Three types of variograms : Linear, with nugget effect, with pure nugget effect.}
\label{fig:trois-types} 
\end{figure}

In this section, the space is always the line 
$\mathbb{R}^1$, and the regionalized scalar variable $y$ is supposed to be known at $n+1$ sample points 
$\{x_{i}$,$\:0\leq i\leq n\}$. The calculations bring us to the three cases  \textit{Linear, Linear with nugget effect and large grid}. The figure \ref{fig:trois-types} illustrates the variogram for these cases. The first
is modeled by a linear variogram of type 
\[
\gamma(x-x')=C\mid x-x'\mid.
\]
A nugget effect appears in the second, and the third case finally
is purely random, without spatial correlation. The second case
also of the linear model, but with a jump to the origin
which is called the nugget effect. This additional variance testifies
regionalization on a small scale, or possibly errors
of measurements because it shows that the mean quadratic increase
between two points, even very close, does not tend to zero. The third
case describes a purely random situation, without spatial correlation.

The first two cases are formulated in terms of variograms, relative
a random function $ y $ only increments are defined,
and stationary. For the third case, that of the large grids,
where covariance and variograms are equivalent, we formulate the
results in terms of covariances. The estimators obtained are always
linear functions of the values $y_{i}=y(x_{i})$ whose coefficients are estimated 
by minimizing $E(y-y^{*})^{2}$. For the three models studied below, 
we express at first the kriging of the residue. In a second step, we give 
ourselves a drift of the form
$a_{1}x+a_{2}x^{2}$ for the first two cases, and $a_{0}+a_{1}x+a_{2}x^{2}$
for large grids, and we estimate the parameters of the drift.
Thirdly, we calculate kriging at the point $x_{0}$
by the additivity theorem. All estimators are given with
their estimation variances. Finally, if we assume linear drift,
i.e. if $a_{2}=0$, the estimator $a_{1}^{*}$ of $a_{1}$ is therefore
the same that was found for the quadratic drift.

\subsection{Linear Variogram}

\paragraph{Kriging the residue $z(x_{0})$}

Let consider in $\mathbb{R}^1$,
the random function representing residue $z$, the variogram $\gamma(x-x')=\gamma(h)=|h|$,
where $|h|$ is the distance between $x$ and $x'$ . We suppose a known
realization of $z$ at two points $x_{i}$ and $x_{i+1}$ and
in a certain number of points outside the interval $[x_{i},x_{i+1}]$.
We want to krige the value $z(x)$ at point $x_{0}=\varepsilon x_{i+1}+(1-\varepsilon)x_{i}$
comprised between $x_{i}$ and $x_{i+1}$. A
solution of type $\nu z_{i+1}+(1-\nu)z_{i}$. The kriging at $x_{0}$
is thus given by: 
\begin{equation}
z^{*}(x_{0})=\varepsilon z(x_{i+1})+(1-\varepsilon)z(x_{i})\,,\qquad(x_{i}\leq x_{0}\leq x_{i+1})
\label{eq:krigeage_linaire}
\end{equation}
which only involves the residues at the two abscissae that frame
$x_{0}$. The Lagrange factor is equal to $0$, and the calculation of
the variance of kriging $\sigma_{K}^{2}$ gives: 
\[
\sigma_{K}^{2}=2\frac{(x_{i+1}-x_{0})(x_{0}-x_{i})}{x_{i+1}-x_{i}}=
2\varepsilon(1-\varepsilon)(x_{i+1}-x_{i})\,.
\]

\paragraph{Estimation of the drift}

Given $x_{1},x_{1},...x_{N}$ $N$ be points at arbitrary distances 
from each other. We search for an optimal estimator $a_{1}^{*}x+a_{2}^{*}x^{2}$
with a shift $a_{1}x+a_{2}x^{2}$ when the variogram is $\gamma(h)=\mid h\mid$.
Let us define: 
\begin{equation}
p_{0}=\frac{x_{2}-x_{1}}{2}\:\quad p_{n}=\frac{x_{N}-x_{N-1}}{2}\;\quad p_{i}=\frac{x_{i+1}-x_{i-1}}{2}\,\qquad(1<i<n)\label{eq:poids}
\end{equation}

It is convenient to take the point  $\frac{1}{2}(x_{1}+x_{N})$ as origin,
which is the origin adapted to local neighborhoods, and to define
\[
c_{2}=\frac{1}{\frac{1}{2}(x_{1}^{2}+x_{N}^{2})(x_{N}-x_{1})-\sum p_{i}x_{i}^{2}}
\]

The expressions of $a_{1}^{*}$ and of $a_{2}^{*}$ are therefore 
\[
a_{1}^{*}=\frac{y_{N}-y_{1}}{x_{N}-x_{1}}
\]
\[
a_{2}^{*}=c_{2}[\frac{(x_{N}-x_{1})(y_{N}+y_{1)}}{2}-\sum p_{i}y_{i}]
\]

\paragraph{Kriging at point $x_{0}$}

By application of the additivity theorem, the expression of kriging
at the point $x_{0}$ is obtained by replacing, in (\ref{eq:krigeage_linaire}),
$z(x_{i})$ and $z(x_{i+1})$ by their estimations: 
\[
y^{*}(x_{0})=\varepsilon[y(x_{i+1})-a_{1}^{*}x_{i+1}-a_{2}^{*}x_{i+1}^{2}]+(1-\varepsilon)[y(x_{i})-a_{1}^{*}x_{i}-a_{2}^{*}x_{i}^{2}]+a_{1}^{*}x_{0}+a_{2}^{*}x_{0}^{2},
\]
given, after simplifications: 
\[
y^{*}(x_{0})=\varepsilon y(x_{i+1})+(1-\varepsilon)y(x_{i})-a_{2}^{*}\varepsilon(1-\varepsilon)(x_{i+1}-x_{i})^{2},
\]
The coefficient $a_{1}^{*}$ of the term at $x$ of the drift disappears,
and the variance $\sigma_{U}^{2}$ de $y^{*}(x)$ is the sum of the variances
of kriging of the residue and the drift: 

\[
\sigma_{U}^{2}=\sigma_{K}^{2}+\sigma_{D}^{2}=2\varepsilon(1-\varepsilon)(x_{2}-x_{1})+c_{2}\left[\varepsilon(1-\varepsilon)(x_{i+1}-x_{i})^{2}\right]^{2}
\]

When the $x_{i}$ abscissa is regular and distant $h$ between
they, the term $c_{2}$ is simplified in
\[
\frac{6}{N(N^{2}-1)}\;\frac{1}{h^{3}},
\]
and the kriging of $x_{0}$ becomes
\[
y^{*}(x_{0})=\varepsilon y(x_{i+1})+(1-\varepsilon)y(x_{i})-a_{2}^{*}h^{2}\varepsilon(1-\varepsilon),
\]
of variance

\[
\sigma_{U}^{2}=2\varepsilon(1-\varepsilon)h+\frac{6\varepsilon^{2}(1-\varepsilon)^{2}}{N(N^{2}-1)}h.
\]

\paragraph{Example}

\begin{figure}
\begin{centering}
\includegraphics[scale=0.5]{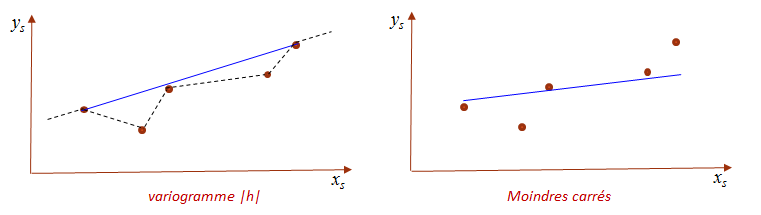} 
\par\end{centering}
\caption{Two models for the same data}
\label{vario-lin} 
\end{figure}

The figure \ref{vario-lin} represents the same data, which are modeled on the left, by a linear variogram and a linear drift, and on the right by a pure nugget effect. The drift is represented in lines
full. On the left, it connects the two extreme points, and on the right
it minimizes the squares of the vertical projections of the experimental points
on her. The kriging is indicated on the left by dashed lines:
it is formed of straight segments connecting the experimental points
neighbors. Extrapolation beyond the extreme points prolongs the drift.
On the right, kriging and drift coincide, and are obtained by the estimator
least squares.

\subsection{Linear Variogram with nugget effect}

\paragraph{Kriging the residue $z(x_{0})$}

For the kriging of the residual we consider $N$ samples
regularly distributed in $\mathbb{R}^1$.
The system to be solved (\ref{eq:system_ku_vario}) for the variograms:

\[
\left\{ \begin{array}{l}
\gamma(i,j)=\gamma(x_{i}-x_{j})=|x_{i}-x_{j}|=|h|\qquad\textrm{ si \ensuremath{x_{i}\neq x_{j}}}\\[2ex]
\gamma(x_{i}-x_{j})=-C\qquad\;\;\qquad\textrm{ si \ensuremath{x_{i}=x_{j}}}.
\end{array}\right.
\]

The simplest method here is to gather the data in
three groups, and take into account only the two-point values
which enclose $x_{0}$ plus the mean $m$ of the samples. We have
then a system of three equations with three unknowns easy to solve.
When the nugget effect tends to zero, we find the linear case
previous, and when it becomes very large the case of large grids
presented a little further.

\paragraph{Estimation of the drift}

For the calculation of the drift we will assume the data to be regularly
spaced. The nugget effect makes this assumption acceptable, and it
can be used as an approximation for irregular data.
$y$ is supposed to be known at $n+1$ points $x$ abscissas:
\[
x_{i}=(i-\frac{n}{2})h
\]
(originated in the data center). The variogram is the sum of one
linear term and variance nugget effect $C$ : 
\[
\gamma(x_{i}-x_{j})=\mid i-j\mid,\:\:x_{i}\neq x_{j}\qquad\quad\gamma(x_{i}-x_{j})=-C,\:\:x_{i}=x_{j}
\]

\paragraph{Linear Drift}

Consider first the case where the drift is reduced to the term $a_{1}x$
and where we estimate the factor $a_{1}$ par $a_{1}^{*}$. Let $\alpha$
and $\beta=\alpha^{-1}$ the two roots of the equation
\[
\alpha^{2}-2(1+\frac{h}{C})\alpha+1=0
\]
the estimator $a_{1}^{*}$ is 
\[
a_{1}^{*}=\sum_{i}\lambda_{1}^{i}y(x_{i})
\]
with
\[
\lambda_{1}^{i}=\frac{\alpha^{i-n/2}-\beta^{i-n/2}}{2Ch}
\]
and the variance of $a_{1}^{*}$ evaluates to
\[
\sigma_{D}^{2}=\frac{\alpha^{1+n/2}+\alpha^{-n/2}}{(1-\alpha)Ch}
\]

\paragraph{Quadratic Drift}

If now the drift has the form $a_{1}x+a_{2}x^{2}$, the 
optimal estimator is $a_{1}^{*}x+a_{2}^{*}x^{2}$ ,the term $a_{1}^{*}$ is
the same as below and $a_{2}^{*}$ is written as: 

\[
a_{2}^{*}=\sum_{i}\lambda_{2}^{i}y(x_{i})
\]
with $\lambda_{2}^{i}$ of the form
\[
\lambda_{2}^{i}=f(\alpha^{i-n/2}+\beta^{i-n/2})+g
\]
where
\[
f=\frac{1}{2h^{2}(E-\frac{n(n+2)}{12}B)}\qquad g=-\frac{2B}{n+1}\:f
\]
with
\[
B=\sum_{i}\alpha^{i-n/2}\qquad E=\sum_{i}(n/2-i)^{2}\alpha^{i-n/2}
\]
The variance of $a_{2}^{*}$ is
\[
D^{2}(a_{2}^{*})=\frac{g}{h}
\]

\subsection{Large grids and least squares}
As previously, we know 
the values of $y$ at $N$ training samples at points $x_{i}$ 
anywhere on the line, and we want to krige $y$ at point $\mathbf{x}_{0}$
falling within two known points $\mathbf{x}_{i}$ \& $\mathbf{x}_{i+1}$.

\paragraph{Kriging the residues}

The residue $z$ is modeled by a random stationary function with unknown mean, and finite variance $C=\sigma(0,0)$ with small range (the range is the distance beyond which data become independent). In such as case the variogram $\gamma(x_{i}-x_{j})=\gamma_{i,j}$
and the covariance $\sigma(x_{i}-x_{j})=\sigma_{i,j}$ are equivalent notions, we formulate the results below in terms of covariance. 
We supposed that the distances between two samples are greater than the range. This independence between these $z(x_{i}$) simplifies the system of kriging of the residue (\ref{eq:system_residus}), that reduces to 

\[
C\nu^{i}=\sigma(i,0)\quad\textrm{and \ensuremath{\quad C\nu^{i+1}=\sigma(i+1,0),}}
\]
the other coefficients $\nu$ being null. The kriging of the residue
$z(x_{0})$ evaluates consequently to : 
\begin{equation}
z^{*}(x_{0})=\frac{1}{C}\left[\sigma(i,0)z(x_{i})+\sigma(i+1,0)z(x_{i+1})\right].\label{eq:krigeage_z_grandes_mailles}
\end{equation}
This equation means that if the covariance $\sigma(i,0)$ (resp.$\sigma(i+1,0)$)
equals zero, then $z^{*}(x_{0})$ is equal to $z(x_{i+1})$ (resp.$z(x_{i})$),
and if both covariances are zero then  $z(x_{i+1})$ is equal to
$0$, i.e.to the average of the process. The kriging variance becomes, by applying expression (\ref{eq:variance_residus}):
\[
E[z^{*}(\boldsymbol{x_{0}})-z(\boldsymbol{x_{0}})]^{2}=\frac{1}{C}[C^{2}+\nu^{i}\sigma^{2}(i,0))+\nu^{i+1}\sigma^{2}(i+1,0)].
\]

\paragraph{Estimation of the drift}
From the moment where the distances between samples are superior to the 
range, the parameters $a_{0}^{*},a_{1}^{*}$ and $a_{2}^{*}$
of the drift $a_{0}+a_{1}x+a_{2}x^2$ are ones that given by least squares.
This classical method consists in determining the three coefficients from the drift
by minimizing the sum $\sum_{i}[y(x_{i})-m^{*}(x_{i})]^{2}$.
The criteria does not hold thus on the known points $x_{i}$, and only
concerns that of the drift. Reconsidering the weights $p_{i}$ from the 
equations (\ref{eq:poids}) we find that : 
\begin{align*}
a_{0}^{*} & =\frac{9}{4(}\frac{1}{x_{n}-x_{0})}\sum_{0}^{n}p_{i}y_{i}-\frac{15}{4(x_{n}x_{0})^{2}}\frac{1}{(x_{n}-x_{0})}\sum_{0}^{n}p_{i}y_{i}x_{i}^{2}\\
a_{1}^{*} & =\frac{3}{(x_{n}-x_{0})^{2}}\frac{1}{(x_{n}-x_{0})}\sum_{0}^{n}p_{i}y_{i}x_{i}\\
a_{2}^{*} & =\frac{-15}{4(x_{n}x_{0})^{2}}\frac{1}{(x_{n}-x_{0})}\sum_{0}^{n}p_{i}y_{i}+\frac{45}{4(x_{n}-x_{0})^{4}}\frac{1}{(x_{n}-x_{0})}\sum_{0}^{n}p_{i}y_{i}x_{i}^{2}
\end{align*}

The covariance matrix associated to $a_{i}^{*}$, i.e.
\[
\text{cov}_{i,j}=E[a_{i}^{*}a_{j}^{*}]-a_{i}a_{j}
\]
allows for values 
\[
\text{cov}_{1,1}=\frac{6}{5(x_{n}-x_{0})},\quad cov_{1,2}=cov_{2,1}=0,\quad cov_{2,2}=\frac{15}{14(x_{n}-x_{0})^{3}}
\]

The estimator $m^{*}(x_{0})$ of the drift at $x_{0}$ is written as
\[
m^{*}(x_{0})=a_{0}^{*}+a_{1}^{*}(x_{0})+a_{2}^{*}(x_{0})^{2}
\]

If we suppose that the drift is linear, we need to keep only the first
two coefficients which are again optimal estimators in this case.

\paragraph{Universal kriging at a point}

The kriging of $y(x_{0})$ is done as previously by addition of
kriging of the residues $y(x_{i})-m^{*}(x_{i})$ and the estimator of drift: 
\[
y^{*}(x_{0})=\frac{1}{C}\left[\sigma(i,0)[y(x_{i})-m^{*}(x_{i})]+
\sigma(i+1,0)[y(x_{i+1})-m^{*}(x_{i+1})]\right]+m^{*}(x_{0})
\]
The variance of the error $\mid y(x_{0})-y^{*}(x_{0})\mid$ is the sum of the 
variances of the two estimators.

\section{Operational Modes}
\label{sec:Modes-op}
We have just given several solutions for kriging 1-D components 
$y_{s}$ that occur in the zonality equation (\ref{eq:zonality}). 
We know how to pass 1-D kriging coefficients to those of $y(\boldsymbol{x_{0}})$
using relations (\ref{eq:ersatz_poids_residus}) (\ref{eq:mu_j})
(\ref{eq:krigeage_y}). Now we just need to build the 1-D variogram estimators
from experimental data.

\paragraph{Marginal Variograms}

According to zonal decomposition (\ref{eq:zonality}),
the increase of $y$ between the points $\boldsymbol{x=}x_{1},...x_{s,...}x_{d}$
and $\boldsymbol{x'=}x'_{1},...x'_{s,...}x'_{d}$ can be expressed as:
\begin{equation}
y(\boldsymbol{x'})-y(\boldsymbol{x})=\sum_{s=1}^{s=d}w_{s}[y_{s}(x'_{s})-y_{s}(x_{s})]
\label{eq:accroissements zonaux}
\end{equation}
The independence between the increments of pairs of distinct parameters
results in the nullity of their co-variation:
\begin{equation}
E[y_{s}(x'_{s})-y_{s}(x_{s})][y_{s}(x'_{t})-y_{s}(x_{t})]=0\qquad s\neq t
\label{eq:accroissements indpendants}
\end{equation}
where
\begin{equation}
\gamma(\boldsymbol{x'}-\boldsymbol{x})=E[y(\boldsymbol{x'})-y(\boldsymbol{x})]^{2}=
\sum_{s}w^{2}E[y_{s}(x'_{s})-y_{s}(x_{s})]^{2}=\sum_{s}w_{s}^{2}\gamma(x_{s}-x'_{s})
\label{eq:somme_variogrammes}
\end{equation}

The variogram of $y$ in $\mathbb{R}^d$ is the sum of $d$ 1-D variograms corresponding to each
parameter, just like the covariance in equation (\ref{eq:covar_zonal_total}).
The $\gamma_{s}$ are not experimentally accessible but
on the other hand, we can estimate numerically the marginal variograms
$\tilde{\gamma_{s}}$ of the variation of $y$ w.r.t a single parameter:

\[
\tilde{\gamma_{s}}(h_{s})=E[y(\boldsymbol{x+h})-y(\boldsymbol{x})]^{2}
\qquad\boldsymbol{x}=(0,...0,x_{s},0...0)]^{2},\;\boldsymbol{x+h}=(0,...0,x_{s}+h_{s},0...0)
\]

Since the parameters are independent, the influence on
$\tilde{\gamma}(h_{s})$ of different parameters of $s$ 
produces a nugget effect only, which does not distort the rest of the curve. We can therefore approximate the $\gamma_{s}$ by: 

\begin{equation}
\gamma_{s}\simeq\tilde{\gamma_{s}}(h_{s})-K\label{eq:approx_gamma_s}
\end{equation}
where $K$ is the difference of the two nugget effects of $\tilde{\gamma_{s}}$ and $\gamma_{s}$, unknown one
and the other. The approximation
(\ref{eq:approx_gamma_s}), which gives particular information on the alignment
first points and on the range. This is enough to choose between
three models calculated above. Modeling $\gamma_{s}$
each parameter $s$ is otherwise independent of the others.

Once the variogram models are determined, it remains to choose
if we prefer to estimate the drift of each $ y_ {s} $ on the totality
samples $x_{i, s} 1 \leq i \leq N$, or on slipping neighborhoods.

\paragraph{Metric in $\mathbb{R}^d$ and independence of parameters}
The zonality hypothesis is may be inaccurate, but it does not create 
biased estimators $y^{*}$ and $m^{*}$. They are solely less precise, since 
we haven't taken into account certain correlations. We can amend this hypothesis in two ways. We will Distinguish two types of data, following what all  parameters $s$ can express with a common unit, like a cost, or can not ?

In the first case, the axes of $\mathbb{R}^d$
are homogeneous, and justify the usage of the euclidean distance which allows for axes rotations. The classical method consists in replacing the
$d$ parameters with the decorrelated axes after a principal component analysis (PCA).

In the second case, it is preferable to take as distance :
\begin{equation}
h=\sum w_{s}h_{s}\label{eq:distance_somme}
\end{equation}
where the $h_{s}$ are the Euclidean distances in $\mathbb{R}^1$.
The balls are rectangle parallelepipeds, and the rotations make
us move from one axes $s$ to another without introducing an oblique orientation. The PCA is replaced by the search for the axe that is least correlated with the others, and the second lesser correlated and so on.
If $\text{cor}(s,t)$ designates
the coefficient of correlation between the parameters, the degree of independence of the parameter $s$ can be written as 
\begin{equation}
R(s)=d-\sum_{t} \text{cor}(s,t)
\label{eq:indpendance_totale}
\end{equation}

Equation (\ref{eq:indpendance_totale}) allows to categorize the parameters
by decreasing order of independence, and construct a decision tree which orders the regression of $y$ into components the with decreasing significance.

We note that the distance $h$ appears already embedded in equation
(\ref{eq:somme_variogrammes}), since once each axes admits a
a linear variogram, the metric (\ref{eq:distance_somme}) implies
that the variogram in $\mathbb{R}^d$ should be linear as well.

\section{Classification}
The work done for regression can be applied for classification,
where the classes are obtained by simple thresholds of the regressed
$y^{*}$. In the most general case where the classes have $P$ categories
without a common measure, we replace these scalar values 
$y(\boldsymbol{x})$ and $y^{*}(\boldsymbol{x})$ by these vectors
$\boldsymbol{y}(\boldsymbol{x})=\{y^{1}(\boldsymbol{x}),...y^{p}(\boldsymbol{x})...y^{P}(\boldsymbol{x})\}$
and $\boldsymbol{y}^{*}(\boldsymbol{x})$ define the space at $P$
dimensions of the classes, where each vector is of the type $\boldsymbol{y}(\boldsymbol{x})=(0,..,1,...0)$,
i.e. where one ans only one class is labelled at value $1.$

The vector kriging is termed as cokriging and does not limit itself to scalar independent variables \cite{GM_71a,Wackernagel}.
We remark that the variance of $\boldsymbol{y}(\boldsymbol{x})$ always exists. Cokriging takes into account the co-variations of features w.r.t to classes, thus appearing in the covariance matrix
$\sigma^{p,q}(\boldsymbol{x,x+h})=E[y^{p}(\boldsymbol{x}-\boldsymbol{m}^{p}),y^{q}((\boldsymbol{x+h})-\boldsymbol{m}^{q}${]}.

The available data  do not allow for a general estimator of the matrix of
functions $\sigma^{p,q}$. To overcome this trouble the simplest hypothesis consists here in considering the diverse classes evolving in similar manner, and holding the following true:
\begin{equation}
\sigma^{p,q}(\boldsymbol{x,x+h})=K^{p,q}\sigma(\boldsymbol{x,x+h}),
\label{eq:simple-cokrigeage}
\end{equation}
with a strictly positive matrix for the weight $K^{p,q}$ and a covariance function 
$\sigma(\boldsymbol{x,x+h})$. The hypothesis (\ref{eq:simple-cokrigeage})
can be expressed equivalently in terms of variograms. This condition
allows us to estimate the drift and the kriging of each class
at $\boldsymbol{x_{0}}$ independently of the others \cite{GM_71a}.
The hypothesis (\ref{eq:simple-cokrigeage}) is not contradictory with the independence of parameters $s$, since it holds on the output domain and not on the input domain.

The implementation of solutions for classification leads to the same calculation of kriging as that of regression. The weights $K^{p,q}$
are proportional to covariances $\sigma^{p,q}(\boldsymbol{x,x})$
accessible empirically. The kriging at point $\boldsymbol{x_{0}}$
of $\mathbb{R}^d$ yields to $P$ estimators $y^{p*}(\boldsymbol{x_{0}})$ of the
mean of the category $p$ en $\boldsymbol{x_{0}}$. Like $y^{p}$ can only be $1$
or $0$, the quantity $y^{p*}$ is also an estimator of the probability of the
category $p$ at $\boldsymbol{x_{0}}$. We choose finally the most probable
class at $\boldsymbol{x_{0}}$, which is one where the estimator is the closest
to $1$. This approach does not take into account the sum of categories at
$\boldsymbol{x_{0}}$ to be equal to $1$. To use this information we need to
formulate the kriging in the context of \textit{compositional data} as by V.
Pavlowski et J.J. Egozcue. 
\medskip
 
\bibliography{bibliography}
\bibliographystyle{apalike}
\end{document}